# Conditioning Large Language Models on Legal Systems? Detecting Punishable Hate Speech

**Florian Ludwig[1], Torsten Zesch[2], Frederike Zufall[3]**
[1] Central Office for Information Technology in the Security Sector (ZITiS), Germany
[2] Computational Linguistics, CATALPA, FernUniversität in Hagen, Germany
[3] Karlsruhe Institute of Technology (KIT), Germany;
Waseda Institute for Advanced Study, Waseda University, Tokyo, Japan
Florian.Ludwig@ZITiS.bund.de, torsten.zesch@fernuni-hagen.de, zufall@kit.edu

## Abstract

The assessment of legal problems requires the consideration of a specific legal system and its levels of abstraction, from constitutional law to statutory law to case law. The extent to which Large Language Models (LLMs) internalize such legal systems is unknown. In this paper, we propose and investigate different approaches to condition LLMs at different levels of abstraction in legal systems. This paper examines different approaches to conditioning LLMs at multiple levels of abstraction in legal systems to detect potentially punishable hate speech. We focus on the task of classifying whether a specific social media posts falls under the criminal offense of incitement to hatred as prescribed by the German Criminal Code. The results show that there is still a significant performance gap between models and legal experts in the legal assessment of hate speech, regardless of the level of abstraction with which the models were conditioned. Our analysis revealed, that models conditioned on abstract legal knowledge lacked deep task understanding, often contradicting themselves and hallucinating answers, while models using concrete legal knowledge performed reasonably well in identifying relevant target groups, but struggled with classifying target conducts.

## 1 Introduction

A concrete legal problem or question, when assessed through the lens of a trained lawyer, is evaluated in light of a whole legal system with several levels of abstraction standing behind it: from higher-level constitutional principles and fundamental rights to be weighed out to a legal ground in statutory law, down to prior case law. The extent to which Large Language Models (LLMs) internalize such legal knowledge, however, is increasingly debated (Fei et al., 2023; Dahl et al., 2024a; El Hamdani et al., 2024). The question arises with respect to numerous legal problems that may be put forward to generative models in the form of prompts. For traditional supervised machine learning approaches, legal knowledge is implicitly encoded in the annotated data instances. However, these approaches must first reconstruct the actual task from the training data, which proves difficult for such a complex and fine-grained classification task, especially when the number of labeled data samples is limited. Another drawback of these approaches is their limited adaptability to diverse legal systems worldwide, each of which constitutes a complex framework with numerous statutes, dispositions, and case law. Adapting them would require re-annotating the same data multiple times, making the process both impractical and time-consuming.

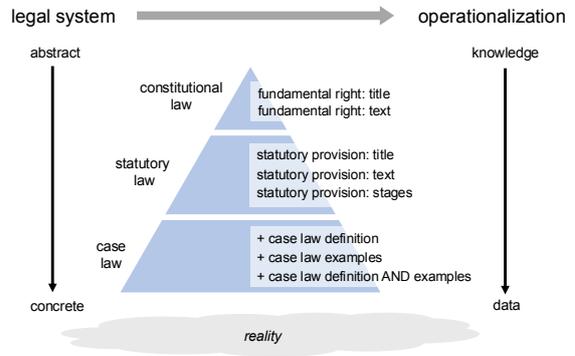

Figure 1: Different levels of abstraction in a legal system and their operationalization.

On the other hand, LLMs can use their inherent knowledge and linguistic skills to generalize to various tasks based on the task descriptions, potentially enabling them to adapt to different legal systems without the need for extensive re-annotation of data. This paper examines the extent to which LLMs can be conditioned on a legal task based on the different levels of abstraction in which it is embedded throughout the legal system. The legal task that we focus on is the question: does a specific text constitute incitement to hatred based on § 130 of the German Criminal Code?

We experiment with end-to-end as well as problem decomposition-based classification approaches, whereby we perform knowledge conditioning on different levels of abstraction from constitutional law, to statutory law and down to case law (see Figure 1). In summary, we make the following contributions:

- Provide an operationalization for conditioning LLMs on a concrete legal task based on the levels of abstraction in legal systems (Section 2).

- Propose conditioning approaches to test at each level of abstraction whether LLMs can apply the respective legal knowledge to examples of potentially punishable hate speech (Sections 2.1, 2.2, 2.3).

- Perform experiments on how LLMs can be conditioned on these tasks (Section 3) and how well these models perform compared to humans.

## 2 Operationalizing legal system conditioning

Behind each concrete legal decision lies a complex legal system with abstract constitutional principles, statutory corpora, down to case law defining ambiguous terms and the concrete decision on the merits. 'Law' in this sense is a hierarchical system, comprising not only knowledge about a number of cases and how they should be decided, but also statutory law and fundamental rights at the constitutional level, spreading over several levels of abstraction (compare the left side of Figure 1). These levels are not only relevant with respect to substantive content of the law, but also relate to comprehensive concepts such as the separation of powers between the legislature, judiciary and administration in modern states.

At the constitutional level, legal systems usually describe general principles for the organization of the state, but also fundamental rights as a value system that permeates the entire legal system. Statutory law set by the legislature usually contains legal notions and open-textured terms that are further concretized and defined by case law. Especially, in criminal law, criminal offenses, as defined in criminal codes, are often only phrased as a single sentence, like *"Whoever, [...] incites hatred against a national, racial, religious group or a group defined by their ethnic origin, against sections of the population [...]"*. Behind this sentence, however, lies a complex assessment the respective human lawyer performs when determining whether a particular behavior in the real-world constitutes a specific criminal offense. It is not only long-standing case law, but also a comparison with similar cases that may qualify as precedents what guides the decision-making process. Understanding legal tasks in their legal systematic context and incorporating these levels of abstraction in an operationalization may not only add to their performance, but also provide higher explainability and justification for the respective computational system. Our key assumption is that if we want to understand the extent to which legal knowledge is contained in existing LLMs or how LLMs might be conditioned on this knowledge, we need to approach legal knowledge as a system of abstraction levels.

Our idea is to provide an operationalization of a concrete legal question together with the underlying normative levels standing behind it throughout a legal system. We assume that hierarchical organization and levels of abstraction in a legal system correspond to artificial intelligence approaches, from explicit knowledge to increasingly data-driven models, from unsupervised LLMs to supervised models that incorporate knowledge in the form of labels (Figure 1). Both concepts, legal systems and increasingly implicit methods of AI attempt to approach a complex reality – down to a concrete legal decision or down to raw data at the lowest levels.

Our use case is the question of whether a particular social media post constitutes illegal content by violating the German criminal law provision on incitement to hatred, § 130(1) of the German Criminal Code.[1] This question can

---
[1]A translation of § 130(1) is provided in the Appendix B.

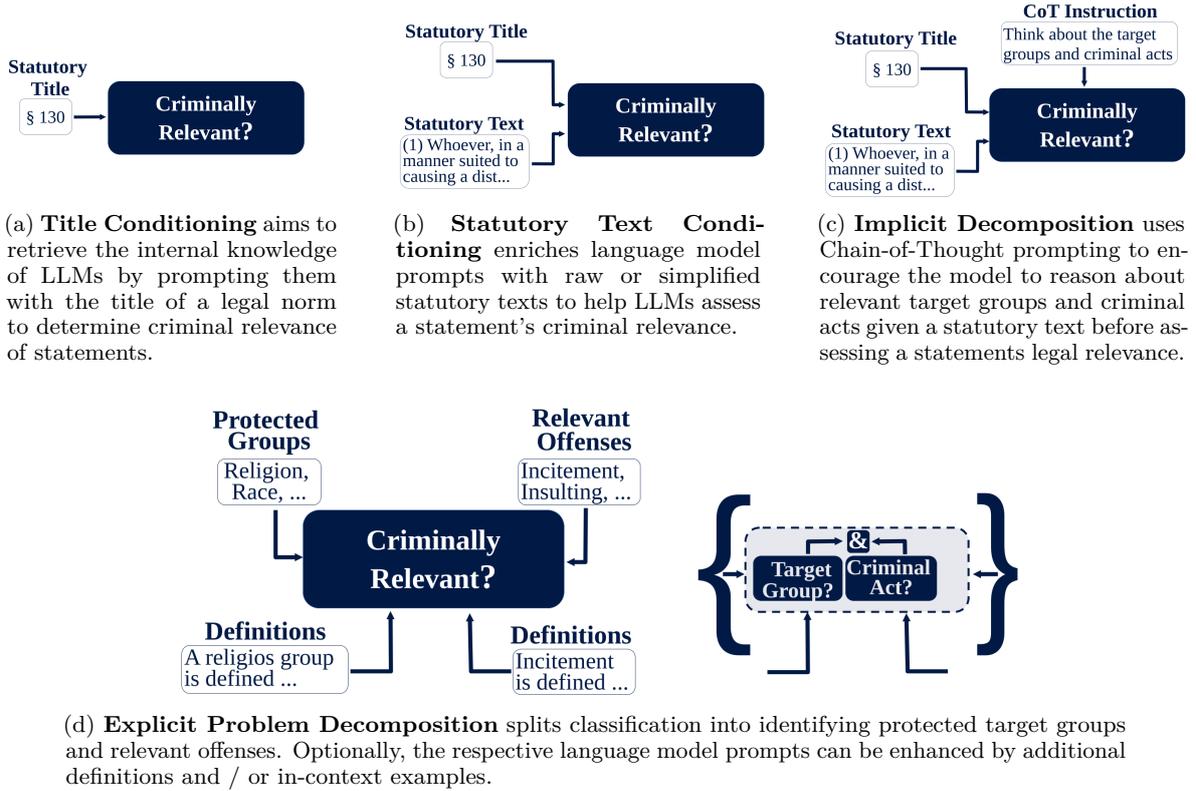

(a) **Title Conditioning** aims to retrieve the internal knowledge of LLMs by prompting them with the title of a legal norm to determine criminal relevance of statements.

(b) **Statutory Text Conditioning** enriches language model prompts with raw or simplified statutory texts to help LLMs assess a statement's criminal relevance.

(c) **Implicit Decomposition** uses Chain-of-Thought prompting to encourage the model to reason about relevant target groups and criminal acts given a statutory text before assessing a statements legal relevance.

(d) **Explicit Problem Decomposition** splits classification into identifying protected target groups and relevant offenses. Optionally, the respective language model prompts can be enhanced by additional definitions and / or in-context examples.

Figure 2: **Approaches** to legal knowledge conditioning, which were investigated in our experiments.

be posed on several levels of abstraction starting from the constitutional level by referring to a protection by fundamental rights (Sec 2.1). Furthermore, the problem may be addressed at the level of statutory law, namely the respective provision of the German Criminal Act, § 130(1) (Sec 2.2). Even further down in the legal system, previous court decisions have as long-standing case law yielded definitions of specific terms in the statutory text (e.g. what constitutes a "group"?). Finally, prior court decisions contain concrete examples on what has been ruled to fulfill certain definitions and constitute punishable conduct (Sec 2.3). In Appendix C, we show examplary prompts for each of the following approaches.

## 2.1 Fundamental rights

Located at the highest abstraction level in legal systems, fundamental rights and their balancing stands behind most statutory provisions and conflicts that are solved in concrete case law. The question of whether a specific social media post constitutes a punishable criminal offense may also be asked from the perspective of fundamental rights. Examples of all prompts are shown in the Appendix C.1.

**Fundamental Rights Title Conditioning** Since all proposed language models were pre-trained with legal data, these models should theoretically be familiar with legal knowledge. The first way to operationalize this dimension is to exploit this internal knowledge by conditioning the models on the title of the constitutional norm. This was done with the following instruction:

> *Is the following text covered by freedom of speech under Article 5(1) GG?*

**Fundamental Rights Text Conditioning** To condition the models on richer context, we provide the models the legal text of Article 5(1) GG in the prompt, which is shown in the Appendix A. After providing the prompt, we instruct the models as follows:

> *Is the following text covered by the freedom of speech according to the above-mentioned article of the Basic Law?*

## 2.2 Statutory law

Going one level of abstraction lower in legal systems (Figure 1), we also perform experi-ments on the level of statutory law. We propose prompts that enable LLMs to determine whether a social media post constitutes punishable incitement to hatred as codified in § 130(1) of the German Criminal Code (StGB). Examples of all proposed prompts are shown in the Appendix C.2.

**Statutory Title Conditioning** In this approach, we directly prompt the models whether a given text is criminally relevant according to a specific legal norm, in our case according to § 130 (1) of the German Criminal Code.

**Statutory Text Conditioning** For this approach, we include the statutory texts, in this case the texts of § 130 (1) of the German Criminal Code (Appendix B), in the prompt and instruct the models to assess the criminal relevance of a given post according to that norm.

**Simplified Statutory Text Conditioning** Legal statutory texts are often written in a highly abstract and complex style, making them difficult to interpret. Therefore, we provide a simplified version of these texts in the prompt. While the simplified texts may not be entirely accurate from a legal perspective, they could still provide a more accessible and useful context for the models.

**High-Level Stages** CoT prompting (Wei et al., 2022) tells the model at a high level how to address the problem based on elements of the legal text, but without the explicit decomposition into stages explained in the following sections. In this approach, we use this technique by first instructing our models to think about the target groups addressed and the target conduct described in the legal text:

> Check whether the statement refers to one of the target groups mentioned and whether it fulfills an offence described above. Specify the relevant target groups and behaviors.

After that, we prompt the model to determine the criminal relevance of the statement.

**Stages** We further decompose the problem into the stages a lawyer would perform when assessing § 130(1) of the German Criminal Code as described by (Zufall et al., 2022). The assessment would, in a first stage, ask whether an utterance targets a *racial*, *national*, *religious* or *ethnic* group, or sections of the *German population*, as listed by the statutory text.

A second stage in the assessment would ask whether one of these protected target groups is subject to a relevant conduct: *inciting hatred*, *calling for violence*, and violating the human dignity through *insulting*, *slandering* or *disparaging*. We therefore prompt our language models to solve these subtasks. A statement is classified as punishable based on § 130(1) of the German Criminal Code if at least one target conduct was committed against a protected group. The prompts are organized in prompt chains, in which the models answers to previous prompts are incorporated into the next prompt, as shown in the Appendix C.2.

### 2.3 Case law

At the lowest level of abstraction in legal systems lies the concrete decision on the case. We operationalize this level by complementing the terms in statutory through existing definitions from long-standing case law. Examples of all prompts are shown in the Appendix C.3.

**Stages + Definitions** Terms in statutory texts are often intentionally kept vague to allow for a adaptation and give discretion to administrative and judicial deciders. These terms are defined through court decisions and may become binding case law over time that lawyers read into the statutory term when applying it. For instance, the term 'incitement to hatred' is defined[2] in a leading decision by the German Federal Court of Justice as follows:

> The incitement to hatred (§ 130(1)) must be objectively suitable [...] increase a heightened hostile attitude against [...] that goes beyond mere rejection and contempt.

The annotation framework of Zufall et al. (2022) provides the definitions of relevant target groups and target conduct, prepared and understandable for legal laypeople. We extend the prompts of the *Stages* approach to include these definitions. We placed the definitions before the instructions in the respective parts of our prompt chains.

**Stages, Definitions + Examples** In addition to definitions, case law also provides precedents in the form of real cases that have already been decided by the courts. For our operationalization these may serve as examples in the prompts to improve the performances of LLMs (Dong et al., 2024). As the annotation framework cited above also contained examples out of existing case law, we extend the prompts of the "*Stages + Definitions*" approach with these examples. For each proposed subtask (see the *Stages* approach), three positive and three negative examples together with their annotations are appended to the prompt. We refer to this approach as "*Stages, Definitions + Examples*". To determine the influence of the data samples themselves, we propose an additional approach, referred to as "*Stages + Examples*", where instructions, definitions and terms are excluded from the prompt and only the in-context samples are used.

---
[2]Shown here translated from German to English

|  |  | F1 | κ |
|---|---|---|---|
| Humans | Experts | .95 | .79 |
|  | Laymen Stages | .76 | .48 |
|  | Laymen Statutory Text | .69 | .31 |
| Baselines | SVM | .61 | .23 |
|  | Random | .51 | .02 |

Table 1: **Classification results**, achieved by different annotators and baseline approaches.

**Target Classification vs Extraction** In the previous approaches, we prompt the language models among others to classify the target groups into predefined categories. In addition to this target classification approach, we propose a target extraction approach. In practice, most instances of hate speech target a limited number of specific groups. According to the United Nations, 75% of hate speech is directed at minorities.[3] A study analyzing over 27.000 hate speech examples found that 97% of posts targeted only 178 distinct groups (Silva et al., 2016). Given that the most frequently affected groups are well known, we prompt LLMs to identify the exact target group within a statement and compare the result against a manually curated list of potential targets (in our case all target groups from the dataset).

## 3 Experimental Setup

In this section, we introduce the dataset, language models, model inference setup, applied in our experiments.

---

[3]https://www.un.org/en/hate-speech/impact-and-prevention/targets-of-hate, February 11, 2025

### 3.1 Dataset

We performed our experiments on the legal hate speech dataset of Zufall et al. (2022). The dataset consists of 1,000 German instances, whereby 850 samples are real world samples that have not been published. We only use the 150 remaining samples, which were created manually to resemble real hate speech posts. The authors of the study proposed a annotation framework that, among others, helps laypersons to assess whether a post is criminally relevant according to § 130 of the German Criminal Code. Each post received annotations from experts and two groups of laypersons: one group using the annotation framework and one group working without it.

Table 1 shows the classification performance in terms of *F1-Score* and *Cohen Kappa agreement* of laypersons with the raw legal text, of laypersons with the annotation framework (*laypeople stages*) and of legal experts on the dataset. While the annotation framework clearly improved the performance of laypeople, there is still a performance gap between laypeople and legal experts, which demonstrate a substantial agreement on the task.

### 3.2 Language Models and Model Inference

We chose the following models for our experiments due to their reasoning capabilities, strong benchmark results and multilingual capabilities: **Phi-3** (Abdin et al., 2024), **Llama 3.1** (Vavekanand and Sam, 2024), **Command R+ (Cmd)** [4] and **Qwen 2** (Yang et al., 2024).

Since prompt variations can significantly impact model performance (Gan and Mori, 2023; Sclar et al., 2023), we averaged performance across six manually crafted paraphrases per classification task, each with different instructions. Provided statutory texts, legal definitions or in-context examples were retained. We applied each prompt to all manually crafted samples from Zufall et al. (2022), first generating free-text responses with the language model. In a second step, we asked the model to format its response in JSON, which allows us to analyze it automatically.[5] We used this two-step approach because prior research showed that enforcing a specific response format can degrade model performance (Tam et al., 2024).

---

[4]https://cohere.com/blog/command-r-plus-microsoft-azure

[5]Minor formatting errors were corrected manually.

|              |                       | Cmd | Llama | Phi | Qwen |
|--------------|-----------------------|-----|-------|-----|------|
| Fund. Rights | Title Art. 5 I GG     | .41 | .11   | .05 | .00  |
|              | Text Art. 5 I GG      | .04 | .09   | .24 | .00  |
| Stat. Law    | Title § 130           | .32 | .05   | .41 | .29  |
|              | Text § 130            | .32 | .00   | .17 | .20  |
|              | Simple Text § 130     | .22 | .00   | .38 | .17  |
|              | High-Level Stages     | .17 | .12   | .09 | .15  |
|              | Stages                | .21 | .00   | .00 | .20  |
| Case Law     | Definitions           | .29 | .16   | .28 | .24  |
|              | Examples              | .09 | .00   | .11 | .01  |
|              | Definitions & Examples| .30 | .39   | .05 | .15  |

Table 2: **Classification performance** of different models across levels of abstractions and approaches. The performance is measured in terms of Cohen Kappa agreement between model predictions and expert annotations.

### 3.3 Baselines

As baseline models, we used a majority classifier and a Support Vector Machine (SVM), which was trained on labeled data. For the SVM, we first computed TF-IDF features and projected them into a lower dimensional space using a Principal Component Analysis. Due to the small amount of labeled data, leave-one-out cross validation was used to determine the performance of the SVM.

## 4 Results and Discussion

Table 2 shows Cohen's Kappa achieved by the proposed approaches across language models. Although no training data is used, some approaches and models are able to outperform the SVM baseline results, which are shown in Table 1. However, there remains a notable gap in performance compared to human experts.

### 4.1 Fundamental rights

In most cases, approaches based on fundamental rights conditioning attained low agreement values (Table 2). Surprisingly, the *Command-R* model achieved its best results using the *Fundamental Rights Title Conditioning* approach despite being conditioned on the most abstract legal knowledge with the least amount of context. Adding legal text did not improve the performances in three out of four cases.

To assess the consistency of the models internal reasoning process, we measured their rate of contradictory decisions. A decision is considered contradictory if a model classifies a statement as either both covered by freedom of expression (section 2.1) and punishable under § 130(1) (section 2.2), or as neither covered by freedom of expression nor punishable under § 130. Both cases are logically mutually exclusive. We computed the ratio of contradictory decisions between the corresponding title-based conditioning approaches and between the corresponding text-based conditioning approaches. For *Command-R*, contradictory decisions occurred in 60% of cases with title-based and 34% with text-based conditioning. For *Qwen 2*, the rates were 27% and 38%, respectively. These high ratios reveal inconsistencies in the internal reasoning process of the models for this task, reducing the trustworthiness of these approaches.

### 4.2 Statutory law

Unexpectedly, within the approaches based on the abstraction level of the statutory law, the *Statutory Title Conditioning* approach outperformed the other approaches in three out of four cases, despite using the least specific legal knowledge. Similarly to the results of section 4.1, it raises the question whether language models understand the task and perform internal reasoning when only the statutory title is provided. On the other hand, providing models with task decomposition knowledge, whether high-level or explicit, did not improve performance. In most cases, it resulted in worse outcomes than previous approaches, despite incorporating additional expert knowledge in the prompts.

Table 3 presents achieved agreement values between model predictions, when prompted with non-existing or irrelevant paragraphs, and expert labels, annotated for § 130(1) of the German Criminal Code. Since the queried paragraphs are either irrelevant or non-existent, we would expect the models to predict the answer "not relevant" every time, thus achieving a random agreement value of roughly 0. However, the correlation values mostly exceeded random levels, particularly with the *Statutory*

| Approach | Model | §120 | §123 | §300 | §324 | §400 | §1000 | §130 |
|---|---|---|---|---|---|---|---|---|
| **Title** | Cmd | .52 | .52 | .46 | .48 | .40 | .41 | .32 |
| | Qwen | .35 | .34 | .28 | .36 | .37 | .36 | .29 |
| **Text** | Cmd | .03 | .07 | .05 | .05 | / | / | .32 |
| | Qwen | .16 | .27 | .27 | .34 | / | / | .29 |
| **Simplified Text** | Cmd | .15 | .19 | .00 | .03 | / | / | .22 |
| | Qwen | .15 | .22 | .23 | .26 | / | / | .17 |

Table 3: Cohen's Kappa agreement achieved through prompting with non-existent or irrelevant paragraphs of the criminal code. The table shows that correlation values mostly exceed random levels, particularly with the *Statutory Title Conditioning* approach.

| Approach | Cmd | Qwen |
|---|---|---|
| Target Classification | .15 | .31 |
| + Definitions | .29 | .32 |
| + Examples | .05 | .00 |
| + Definitions + Examples | .29 | .11 |
| Target Extraction | **.34** | **.51** |

Table 4: **Target group classification** performance in terms of Cohen's Kappa.

*Title Conditioning* approach. Thus, evidence suggests that models do not deeply understand the task when using the *Statutory Title Injection* approach. The models tend to disregard the specific paragraph provided in the prompt and thus hallucinating their answers.

When an additional statutory text (either raw or simplified) was provided, the models showed a reduction in this type of hallucination. This indicates that LLMs are able to understand the broader context of the legal texts and apply it to concrete samples of hate speech. However, since the agreement values remain above random levels and a significant performance gap persists between models and legal experts, the results indicate that the models struggle to apply fine-grained details and statutory terms to these samples. The struggles with understanding legal knowledge and applying it to concrete hate speech samples is in line with the findings of (Luo et al., 2023).

### 4.3 Case law

Case law-based approaches performed best for only one out of four models (Table 2). Overall, the models benefited from the inclusion of case law definitions in the prompt, which mostly outperformed the *Stages* approach, highlighting the importance of defining case law terms for language models. On the other hand, including examples into the prompt had mixed effects, boosting the performance of some models but harming the performance of others. The exclusive use of examples without the use of definitions led to poor results. This can be partly explained by our observation that the models often refused to answer for safety reasons when confronted with additional examples of hate speech.

**Target Group** Table 4 shows Cohen-Kappa agreement values for different approaches to target group classification and extraction. The results show that models performed better when target groups were extracted from the text rather than classified into predefined categories, supporting the findings of (Sicilia et al., 2024) that LLMs excel at information extraction tasks. As with the overall performance of the model, including definitions tend to improve the performances of the target group categorization approach, while including examples provide mixed results. Generally, the average performance for classifying or extracting target groups was higher than most performances of classifying target conduct.

**Target Conduct** Table 5 shows the Cohen Kappa agreement values, achieved for the different approaches and with respect to the target conduct. Only the conduct of *Calling for Violence* is classified with a reasonable performance, while the classification performances for the other tasks were poor. The results are similar to the results of legal laypersons, which achieve significantly higher agreement values for classifying target groups and classifying the conduct of *Calling for Violence* than for classifying the conduct of *Inciting Hatred* (Zufall et al., 2022).

| Conduct | Approach | Cmd | Qwen |
|---|---|---|---|
| Incitement | Stages | .03 | .00 |
| | +Examples | .03 | .01 |
| | +Definitions | .00 | .03 |
| | +Def/+Ex | .05 | .02 |
| Violence | Stages | .31 | .58 |
| | +Examples | .30 | .57 |
| | +Definitions | .52 | .68 |
| | +Def/+Ex | .29 | .42 |
| Insulting | Stages | .03 | .02 |
| | +Examples | .04 | .02 |
| | +Definitions | .03 | .01 |
| | +Def/+Ex | .04 | .01 |
| Slandering | Stages | .02 | .01 |
| | +Examples | .03 | .01 |
| | +Definitions | .06 | .01 |
| | +Def/+Ex | .02 | .01 |
| Disparaging | Stages | .09 | .07 |
| | +Examples | .04 | .00 |
| | +Definitions | .10 | .11 |
| | +Def/+Ex | .17 | .07 |

Table 5: **Targeting Conduct classification** performance (Cohen's Kappa Agreement).

## 5 Related Work

Legal knowledge has early on been modeled into legal expert systems as ontologies (Oberle et al., 2012), semantic networks (Branting, 2003) or frames (Rissland and Ashley, 1987; Ashley, 1991). In recent years, research combining NLP methodologies and law had a strong focus on supervised learning tasks, especially legal judgment prediction (Aletras et al., 2016; Chalkidis et al., 2019). Work on judgment prediction experimented with using either the factual section of judgment as input, or combined it with the legal reasons and decision on the merits as training data.

Trautmann et al. (2022) has coined the term of 'legal prompt engineering' with experiments on judgment prediction. However, experiments by Dahl et al. (2024b) found that the hallucination rate of LLMs tested on huge case law corpora was up to 58%, raising doubts to what extent concise legal knowledge is actually contained in LLMs. Jiang and Yang (2023) experimented with chain-of-thought and legal syllogism prompting for legal judgment prediction. Savelka et al. (2023) augmented LLMs with a legal information retrieval module to provide explanation of statutory terms.

How legal knowledge inherent in LLMs may be accessed through engineering suitable prompts has been explored in the area of hate speech detection (Morbidoni et al.; Chiu et al., 2021; He et al., 2023; Han and Tang, 2022; Del Arco et al., 2023). Kumarage et al. (2024) explore three different types of prompting schemes: direct yes or no questions, incorporation of basic description of hate speech, and chain-of-thought prompting.

Similarly to our work, AlKhamissi et al. (2022) investigate how hate speech recognition tasks can be prepared for language models so that these models can solve them. They propose a task decomposition and knowledge infusion approach to fine-tune these models in a few-shot manner. The authors propose to split the task into simpler sub-tasks. Additionally, they aim to incorporate common sense and stereotype knowledge from large knowledge bases into a BART-based model (Lewis, 2019). However, they do not take legal systems into account.

## 6 Conclusion

In this work, we investigated to which extent LLMs can be conditioned based on knowledge taken out of different levels of abstraction in legal systems, focusing on the task of identifying punishable hate speech. We experimented with various approaches to legal system conditioning, from the constitutional level to statutory law down to case law. Results showed a significant performance gap between legal experts and models, regardless of the approach or language model used. Surprisingly, approaches with more concrete legal knowledge performed worse on average than high-level end-to-end classification approaches. Further analysis revealed that models conditioned on abstract legal knowledge lacked deep task understanding, contradicting themselves and hallucinating answers when confronted with fictitious or irrelevant offenses. In contrast, models that rely on concrete legal knowledge, similar to human laypersons, have difficulty classifying certain target conducts, while achieving reasonable results in classifying target groups.

# Limitations

We only experiment with one legal system (represented by the dataset used for evaluation). However, our approach would be directly replicable for other legal systems or other areas of criminal law. As the dataset is publicly available, we cannot rule out data contamination, but the overall low performance level makes that unlikely.

# Ethics Statement

Predicting whether a social media posts consists punishable hate speech interferes with the fundamental right of 'free speech'. It could also be abused by governments to suppress opinions. On the other hand, the criminal offense investigated here also protects minorities from discrimination. Both aspects place our work at the center of a balancing decision between those two interests. Our work is neutral towards any political position, but only reflects existing law at several levels of a legal system. We base our ground truth solely on a legal assessment founded on existing German statutory and case law. We believe that our approach contributes to a better understanding and higher transparency in automated content detection methods than conventional approaches, justifying the potential dangers and abuse of our research.

**Race and Gender** § 130(1) of the German Criminal Act explicitly requires the conduct to be directed against a "national, racial, religious group or a group defined by their ethnic origin, against sections of the population." The offense lists these groups as potential targets in order to protect them and prevent discrimination. Our approach is also suitable to be adapted to other groups protected in other legal systems. In view of the inherent biases LLMs may already contain, our approach provides more transparency and investigates targeted protection in favor of groups subject to discrimination.

## A  Text of Article 5

Translation based on: https://www.gesetze-im-internet.de/englisch_gg/

> (1) Every person shall have the right freely to express and disseminate his opinions in speech, writing and pictures and to inform himself without hindrance from generally accessible sources. Freedom of the press and freedom of reporting by means of broadcasts and films shall be guaranteed. There shall be no censorship.

## B  Text of § 130(1)

Translation based on: https://www.gesetze-im-internet.de/englisch_stgb

> (1) Whoever, in a manner suited to causing a disturbance of the public peace,
>
> 1. incites hatred against a national, racial, religious group or a group defined by their ethnic origin, against sections of the population or individuals on account of their belonging to one of the aforementioned groups or sections of the population, or calls for violent or arbitrary measures against them or
>
> 2. violates the human dignity of others by insulting, maliciously maligning or defaming one of the aforementioned groups, sections of the population or individuals on account of their belonging to one of the aforementioned groups or sections of the population
>
> incurs a penalty of imprisonment for a term of between three months and five years.

## C  Examplary Prompts

In this section, we provide example prompts for the different approaches, corresponding to the different levels of abstractions of the legal system.

### C.1  Fundamental Rights

**Fundamental Rights: Title**
**Instruction:**
*Is the following text covered by freedom of speech under Article 5(1) GG?*
**Text:**
**[STATEMENT]**

**Fundamental Rights: Text**
**Legal Text of Articel 5 GG**
(1) Every person shall have the right freely to express and disseminate his opinions [...]
**Instruction:**
*Is the following text covered by the freedom of speech according to the above-mentioned article of the Basic Law?*
**Text:**
**[STATEMENT]**

## C.2 Statutory Law

---

**Statutory Law: Title**

**Instruction:**
*Is the following text criminally relevant according to § 130 StGB, (1)?*

**Text:**
**[STATEMENT]**

---

**Statutory Law: Text**

**Statutory Text of § 130 StGB:**
(1) Whoever, in a manner suited to causing a disturbance of the public peace, [...]

**Instruction:**
*Is the following text criminally relevant according to the above statutory text?*

**Text:**
**[STATEMENT]**

---

**Statutory Law: Simple Text**

**Statutory Text:**
Anyone who incites hatred against people because they have a certain origin, [...]

**Instruction:**
*Is the following text criminally relevant according to the above statutory text?*

**Text:**
**[STATEMENT]**

---

**Statutory Law: Implicit Stages**

**Statutory Text of § 130 StGB:**
(1) Whoever, in a manner suited to causing a disturbance of the public peace, [...]

**Instruction:**
*Check whether the statement refers to one of the target groups mentioned and whether it fulfills an offence described above. Specify the relevant target groups and behaviors. Is the following text therefore criminally relevant according to the above statutory text?*

**Text:**
**[STATEMENT]**

---

**Statutory Law: Stages**

**Instruction:**
Does the following text refer to a group of people, which is considered to be part of the German population?

**Text:**
**[STATEMENT]**

**[MODEL PREDICTION]**

**Instruction:**
Does the text incite hatred against this group according to the legal definition?

---

**Statutory Law: Target Extraction**

**Instruction:**
Analyze the following statement. Is the post aimed at a specific target group? If so, extract any words, phrases or symbols that could indicate this target group.

**Text:**
**[STATEMENT]**

**[MODEL PREDICTION]**

**Instruction:**
Determine the target groups of the statement. If the target group is not clear, please mark the target group as "Unclear"

## C.3 Case Law

**Statutory Law: Stages + Definitions**

**Definition: Part of German Population**
A group is considered to be part of the German population, if [...]

**Instruction:**
Does the following text refer to a group of people, which is considered to be part of the German population?

**Text:**
[**STATEMENT**]

[**MODEL PREDICTION**]

**Definition: Incitement to hatred**
Incitement to hatred must be objectively suitable [...]

**Instruction:**
Does the text incite hatred against this group according to the definition?

[**MODEL PREDICTION**]

**Statutory Law: Stages + Definitions + Examples**

**Definition: Part of German Population**
A group is considered to be part of the German population, if [...]

**Instruction:**
Does the following text refer to a group of people, which is considered to be part of the German population?

**Text:**
[**EXAMPLE STATEMENT 1**]

[**EXAMPLE PREDICTION 1**]

**Text:**
[**EXAMPLE STATEMENT 2**]

[**EXAMPLE PREDICTION 2**]

...

**Text:**
[**STATEMENT**]

[**MODEL PREDICTION**]

**Case Law: Stages + Examples**

**Text:**
[**EXAMPLE STATEMENT 1**]

[**EXAMPLE PREDICTION 1**]

**Text:**
[**EXAMPLE STATEMENT 2**]

[**EXAMPLE PREDICTION 2**]

...

**Text:**
[**STATEMENT**]

[**MODEL PREDICTION**]